\documentclass[10pt,twocolumn,letterpaper]{article}

\usepackage{iccv}
\usepackage{times}
\usepackage{epsfig}
\usepackage{graphicx}
\usepackage{amsmath}
\usepackage{amssymb}
\usepackage{algorithm}
\usepackage{algpseudocode}
\usepackage{tabularx}
\usepackage{threeparttable}
\usepackage{booktabs}
\usepackage{multirow} 
\usepackage{balance}

\usepackage[accsupp]{axessibility}

\usepackage[pagebackref=true,breaklinks=true,letterpaper=true,colorlinks,bookmarks=false]{hyperref}

\iccvfinalcopy 

\def\iccvPaperID{****} 

\ificcvfinal\fi

\begin{document}

\title{Paint Transformer: Feed Forward Neural Painting with Stroke Prediction}

\author{Songhua Liu$^{1,2,*,\dag}$, Tianwei Lin$^{1,*}$, Dongliang He$^1$, Fu Li$^1$,\\Ruifeng Deng$^1$, Xin Li$^1$, Errui Ding$^1$, Hao Wang$^3$\\
$^1$Department of Computer Vision Technology (VIS), Baidu Inc.,\\
$^2$Nanjing University, $^3$Rutgers University\\
{\tt\small $^1$\{liusonghua,lintianwei01,hedongliang01,lifu,dengruifeng,lixin41,dingerrui\}@baidu.com},\\
{\tt\small $^2$songhua.liu@smail.nju.edu.cn, $^3$hw488@cs.rutgers.edu}\\
}

\twocolumn[{%
\maketitle
\begin{figure}[H]
\hsize=\textwidth 
\centering
\vspace{-1cm}
\includegraphics[width=16.7cm]{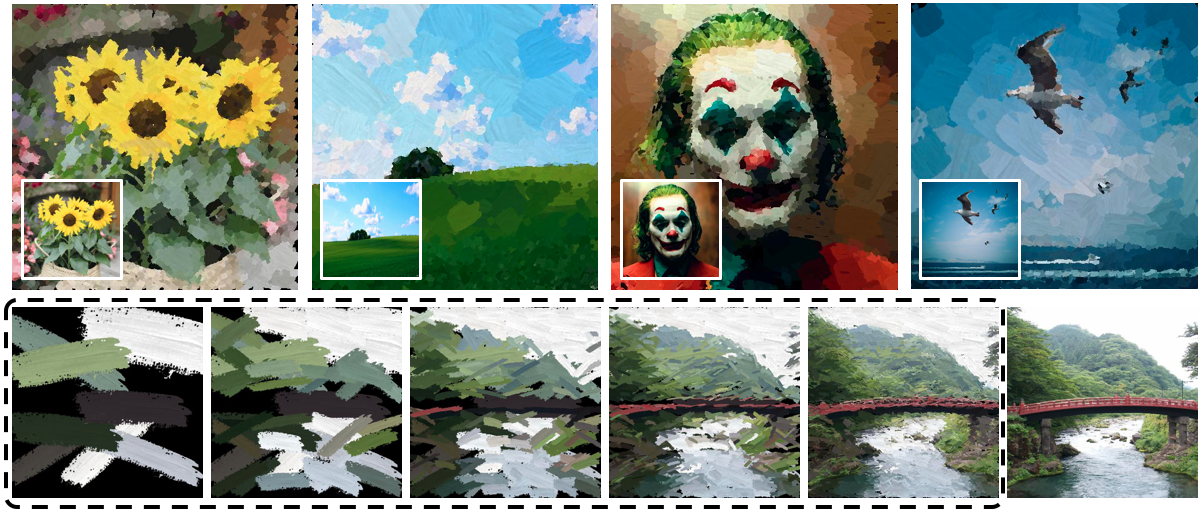}
\caption{Illustration of our results. The second row demonstrates the progressive painting process. Zoom-in for better view.}
\label{fig:overview}
\end{figure}
}]

\renewcommand{\thefootnote}{\fnsymbol{footnote}}
\footnotetext[1]{Equal contribution.}
\footnotetext[2]{This work was done when Songhua Liu was an intern at VIS, Baidu.}
\renewcommand{\thefootnote}{\arabic{footnote}}

\ificcvfinal\thispagestyle{empty}\fi

\begin{abstract}
   Neural painting refers to the procedure of producing a series of strokes for a given image and non-photo-realistically recreating it using neural networks. 
   While reinforcement learning (RL) based agents can generate a stroke sequence step by step for this task, it is not easy to train a stable RL agent. 
   On the other hand, stroke optimization methods search for a set of stroke parameters iteratively in a large search space; such low efficiency significantly limits their prevalence and practicality. 
   Different from previous methods, in this paper, we formulate the task as a set prediction problem and propose a novel Transformer-based framework, dubbed \textit{Paint Transformer}, to predict the parameters of a stroke set with a feed forward network.
   This way, our model can generate a set of strokes in parallel and obtain the final painting of size $512\times 512$ in near real time. 
   More importantly, since there is no dataset available for training the Paint Transformer, we devise a self-training pipeline such that it can be trained without any off-the-shelf dataset while still achieving excellent generalization capability.
   Experiments demonstrate that our method achieves better painting performance than previous ones with cheaper training and inference costs. Codes and models are available\footnote{\href{https://github.com/wzmsltw/PaintTransformer}{PaddlePaddle Implementation}.}.
\end{abstract}

\section{Introduction}

Since ancient times, painting has been a fantastic way for human beings to record what they perceive or even how they imagine about the world. Painting has long been known to require professional knowledge/skills and is not easy for ordinary people. Computer-aided art creation largely fills this gap and enables many of us to create our own artistic compositions. Especially with the coming of AI era, natural images can be transformed to be artistic via image style transfer \cite{li2018learning, huang2017arbitrary, johnson2016perceptual, park2019arbitrary, kolkin2019style} or image-to-image translation \cite{CycleGAN2017, wang2020learning, chen2018cartoongan, yi2019apdrawinggan, yi2020unpaired}. 
%
%
These previous methods typically formulate image creation as an optimization process in the pixel space \cite{gatys2016image} or a feed-forward pixel-wise image mapping with neural networks \cite{isola2017image,CycleGAN2017}. 
Nevertheless, different from pixel-wise operations of neural networks, humans create paintings through a stroke-by-stroke procedure, using brushes from coarse to fine. It is of great potential to make machines imitate such a stroke-by-stroke process to generate more authentic and human-creation-like paintings. Besides, it also has the additional benefit of interpreting how a painting can be created step by step, which might be valuable as a teaching tool. 
Thus, as an emerging research topic, stroke based neural painting is explored to generate a series of strokes for imitating the way that artistic works are created by human painters. Hopefully, with such techniques, the generated paintings can look more like real human created paintings such as oil paint or watercolor.

Generating stroke sequences for painting process is a challenging task even for skilled human painters, especially when the targets have complex compositions and rich textures.
To achieve this goal, some previous works tackle this problem by a sequential process of generating strokes one by one, such as recurrent neural networks (RNN) \cite{zheng2018strokenet,ha2017neural}, step-wise greedy search \cite{haeberli1990paint,litwinowicz1997processing}, and reinforcement learning (RL) \cite{ganin2018synthesizing,zhou2018learning,xie2013artist,huang2019learning,nakano2019neural}.
There are also methods \cite{zou2020stylized,kotovenko_cvpr_2021} tackling this problem via stroke parameter searching using an iterative optimization process.
Although attractive painting results are generated by these methods, there still exists large room for improvement on both efficiency and effectiveness. Sequence-based methods such as RL are relatively fast in inference but suffer from long training time as well as unstable agents. Meanwhile, optimization-based methods \cite{zou2020stylized,kotovenko_cvpr_2021} do not need training, but its optimization process is extremely time consuming. 
These inconveniences motivate us to explore more efficient and elegant solutions for stroke-based painting generation.
Instead of stoke sequence generation, we re-formulate the neural painting task as a feed-forward \emph{stroke set prediction} problem. 
Given an initial canvas and a target natural image, our model predicts a set of strokes and then renders them on the initial canvas to minimize the difference between the rendered image and the target one. This procedure is repeated at $K$ coarse-to-fine scales. At each scale, its initial canvas is the output of the previous scale. As shown in Fig.~\ref{fig:overview}, high-quality final paintings can be generated.


Therefore, the core problem of our method is to train a robust stroke set predictor. 
Interestingly, object detection is also a typical set prediction problem. We are therefore inspired by recent object detector DETR \cite{carion2020endtoend} and propose our novel \textit{Paint Transformer} to generate painting via predicting parameters of multiple strokes with a feed forward Transformer.
However, different from object detection, no annotated data is available for training a stroke predictor.
To overcome such difficulty, we propose a novel self-training pipeline which utilizes synthesized stroke images. 
Specifically, we first synthesize a background canvas image with some randomly sampled strokes; then, we randomly sample a foreground stroke set, and render them on canvas image to derive a target image.
Thus, the training objective of the stroke predictor is to predict the foreground stroke set and minimize the differences between the synthesized canvas image and the target image, where the optimization is conducted on both stroke level and pixel level.
%
%
Impressively, our self-trained \emph{Paint Transformer} shows great generalization capability and can work for arbitrary natural images once trained.
Extensive experiments demonstrate that our feed-forward method can generate paintings with better quality at lower cost compared to existing methods.
Our contributions can be summarized as:

\begin{itemize}
    \item We view stroke-based neural painting problem from an innovative perspective of feed-forward stroke set prediction, instead of stroke sequence generation or optimization-based stroke search. 
    \item A novel Paint Transformer tailored for this task is proposed with a creative self-training strategy to make it well trained without any off-the-shelf dataset. 
    \item Extensive experiments are conducted to validate our approach and demonstrate that state-of-the-art visual quality is achieved, while maintaining high efficiency. 
\end{itemize}

\begin{figure*}
\begin{center}
\includegraphics[width=0.99\textwidth]{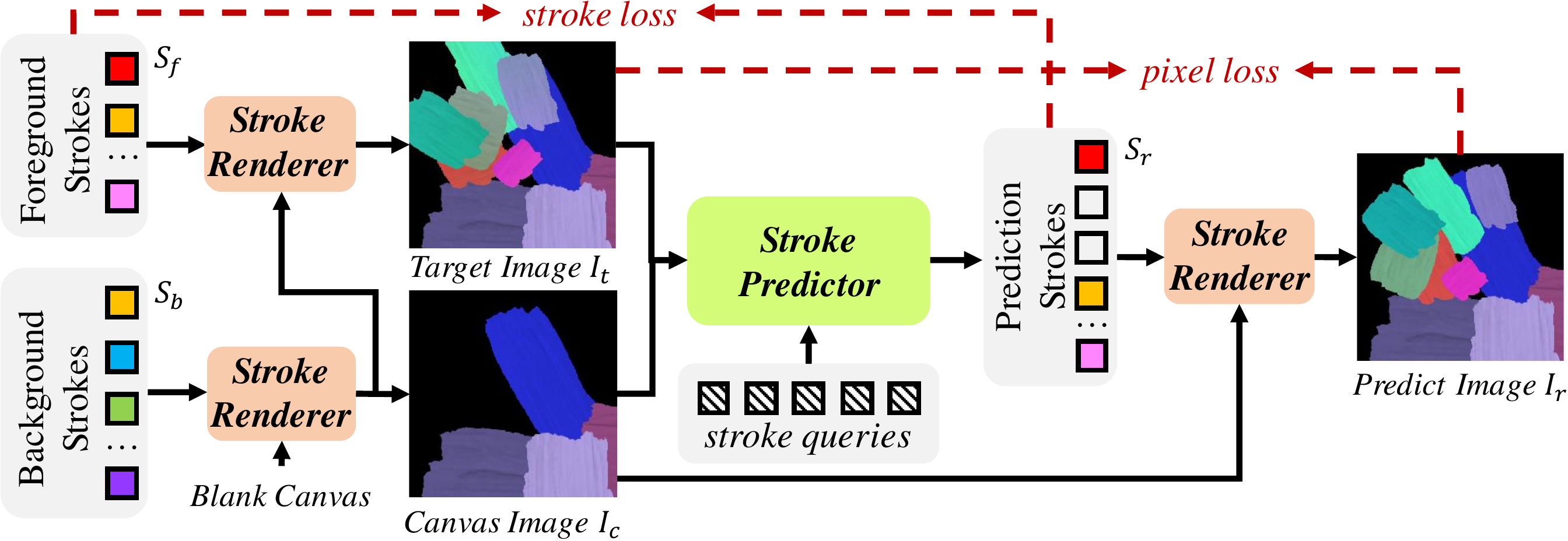}
\end{center}
   \caption{Demonstration of our proposed \emph{self-training} pipeline for Painter Transformer.}
\label{fig:framework}
\vspace{-0.2cm}
\end{figure*}

\section{Related Works}
\subsection{Stroke Based Painting}
It is not a totally new research topic to teach machines how to paint.
Traditional methods usually devise heuristic painting strategies~\cite{hertzmann1998painterly} or greedily select a stroke that minimizes difference from the target image step by step~\cite{haeberli1990paint,litwinowicz1997processing}.
In recent years, RNN and RL are largely applied to generate strokes in a sequential manner.
Ha \etal~\cite{ha2017neural} proposed an RNN-based solution to generate strokes for sketches. 
Ganin \etal~\cite{ganin2018synthesizing} and Zhou \etal~\cite{zhou2018learning} introduced RL for the sketch synthesis task.
These works focus on sketches only, while in \cite{xie2013artist} RL-based strokes generation is explored for ink painting.
By leveraging strengths of CNN, RNN, GAN, and RL, \cite{huang2019learning} provided solutions to generate more photo-realistic paintings.
Nevertheless, training a stable RL agent is difficult due to the alternate and adversarial updates of actors, critics, and discriminators. 
Recently, Zou \etal~\cite{zou2020stylized} proposed a stroke optimization strategy that iteratively searches optimal parameters for each stroke and is possible to be optimized jointly with neural style transfer. 
Similar idea is also adopted in Kotovenko \etal~\cite{kotovenko_cvpr_2021}. 
Although its artistic painting effect is satisfactory, its computational cost largely limits its applicability.
Differently, we formulate neural painting as a problem of feed-forward stroke set prediction, in order to seek better trade-off between performance and efficiency.

\subsection{Object Detection}
Our Paint Transformer is essentially a set prediction model and is largely inspired by object detection. 
Pioneering deep object detection models use an inconvenient two-stage pipeline~\cite{ren2015faster}. 
There are also one-stage object detectors proposed, such as \cite{yolo,redmon2016look,tian2019fcos}.
However, its heavy dependence on post-processing steps such as non-max suppression can still bring much inconvenience. 
Recently, \textit{DETR} \cite{carion2020endtoend} employs Transformer \cite{vaswani2017attention} to produce detection results end-to-end and we find \textit{DETR} quite suitable for our stroke prediction task, since it can perform set prediction without any tricky post-processing. Nevertheless, instead of directly adopting \textit{DETR}, we add binary neurons to predict a stroke should be kept or not. Besides, our model takes two images (current canvas and target images) as input.

\section{Methods}


\subsection{Overall Framework}

We formulate the neural painting as a progressive stroke prediction process. At each step, we predict multiple strokes in parallel to minimize the difference between current canvas and our target image in a feed-forward fashion.
Our Paint Transformer consists of two modules: Stroke Predictor and Stroke Renderer.
As illustrated in Fig.~\ref{fig:framework},
given a target image $I_t$ and an intermediate canvas image $I_c$, \emph{Stroke Predictor} generates a set of parameters to determine current stroke set $S_r$.
Then, \emph{Stroke Renderer} generates the stroke image for each stroke in $S_r$ and plots them onto the canvas $I_c$, producing the resulting image $I_r$. We can formulate this process as:

\begin{equation}
I_r = PaintTransformer(I_c, I_t)
\end{equation}


In Paint Transformer, only Stroke Predictor contains trainable parameters, while Stroke Renderer is a parameter-free and differentiable module. 
To train a Stroke Predictor, as shown in Fig.~\ref{fig:framework}, we propose a novel \emph{self-training} pipeline which utilizes randomly synthesized strokes. 
In each iteration during training, we first randomly sample a foreground stroke set $S_f$ and a background stroke set $S_b$. We then generate a canvas image $I_c$ using Stroke Renderer taking as input $S_b$ and produce a target image $I_t$ by rendering $S_f$ onto $I_c$. Lastly, taking $I_c$ and $I_t$ as input, Stroke Predictor can predict a stroke set $S_r$, after which Stroke Renderer can generate a predicted image $I_r$ taking $S_r$ and $I_c$ as input. 
In other words, Stroke Predictor is trained under a \emph{stroke-image-stroke-image} pipeline, where the optimization is conducted on both stroke level and pixel level.
Therefore, the training objective for the Stroke Predictor is:

\begin{equation}
\mathcal{L} = \mathcal{L}_{stroke}(S_r, S_f) + \mathcal{L}_{pixel}(I_r, I_t),
\end{equation}
where $\mathcal{L}_{stroke}$ and $\mathcal{L}_{pixel}$ are stroke loss and pixel loss separately.
Note that strokes used for supervision are randomly synthesized so that we can generate infinite data for training and do not rely on any off-the-shelf dataset.
Appealing results can be produced by our self-trained Paint Transformer. We will provide detailed description for each part of our method in the following.



\subsection{Stroke Definition and Renderer}

\begin{figure}[t]
\begin{center}
\includegraphics[width=0.9\linewidth]{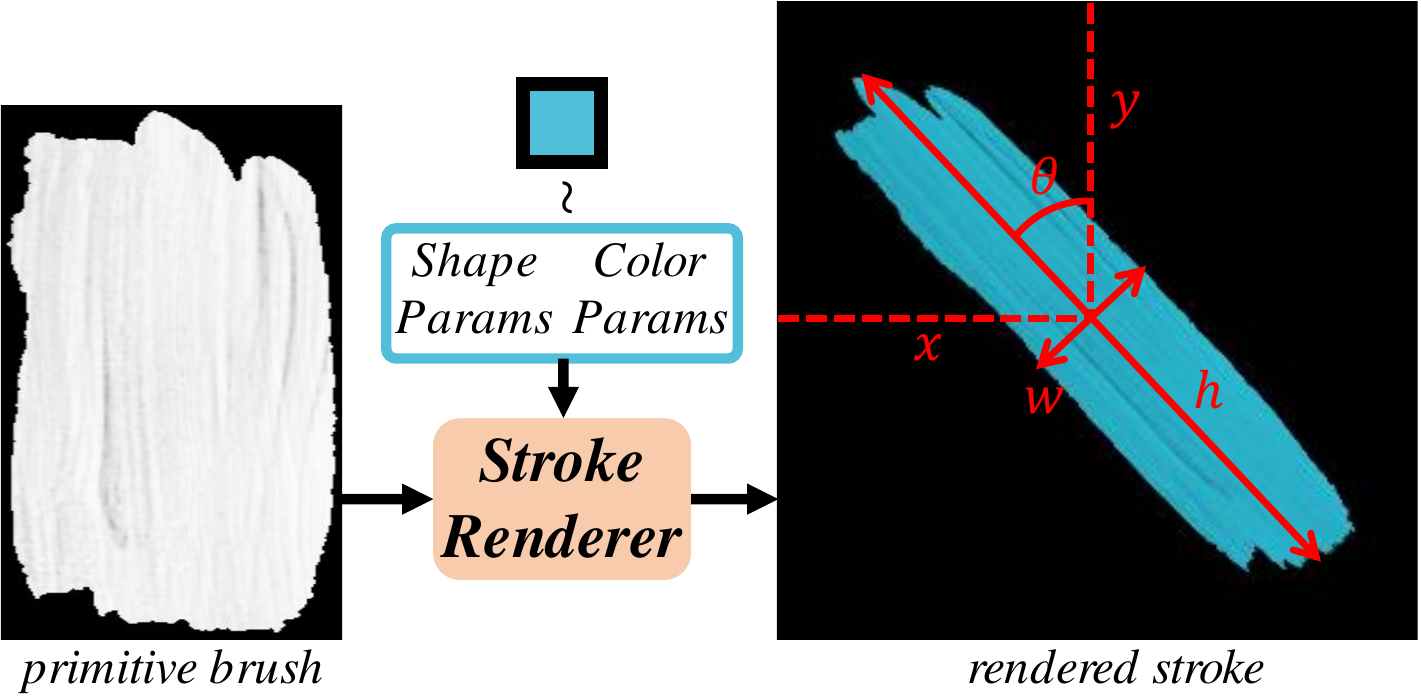}
\end{center}
   \caption{Illustration of Stroke Renderer and parameter definition of a stroke.}
\label{fig:style_render}
\vspace{-0.4cm}
\end{figure}

In this work, we mainly consider straight line stroke, which can be represented by shape parameters and color parameters.
As shown in Fig.~\ref{fig:style_render}, \emph{shape parameters} of a stroke include: center point coordinate $x$ and $y$, height $h$, width $w$ and rotation angle $\theta$.
\emph{Color parameters} of a stroke include RGB values denoted as $r$, $g$ and $b$.
Thus, a stroke $s$ can be denoted as $\left \{x, y, h, w, \theta, r, g, b  \right \}$.

In the task of neural painting, differentiable rendering is one important problem to synthesize stroke images based on stroke parameters and thereby enable end-to-end training of Stroke Predictor. 
Recently, deep neural networks have been widely utilized as differentiable renderers as discussed in \cite{kato2020differentiable}. 
Nevertheless, for the specific stroke definition in this paper, instead of adopting neural networks, we consider a geometric transformation based Stroke Renderer, which does not need training and is differentiable as expected. We denote this Stroke Renderer as:
%

\begin{equation}
I_{out} = StrokeRenderer(I_{in}, S),
\end{equation}
where $I_{in}$ and $I_{out}$ are input and output canvas separately and $S = \left \{  s_i\right \}_{i=1}^n$ is a set of $n$ strokes.
As shown in Fig.~\ref{fig:style_render}, given a primitive brush $I_b$ and a stroke $s_i$, we can modify its color and transfer its shape and location in canvas coordinate system, obtaining its rendered stroke image $\bar{I}_b^i$. 
Meanwhile, we generate a single-channel alpha map $\alpha^i$ with the same shape of $\bar{I}_b^i$ as a binary mask of $s_i$.
Denoting $I^0_{mid} = I_{in}$, we can formulate stroke rendering process as:

\begin{equation}
I^i_{mid} = \alpha^i\cdot \bar{I}_b^i + (1 - \alpha^i)\cdot I_{mid}^{i-1},
\end{equation}
where the output of the Stroke Renderer is $I_{out} = I^n_{mid}$.
Since the whole process can be achieved by linear transformation, Stroke Renderer becomes differentiable.




\subsection{Stroke Predictor}
The goal of our Stroke Predictor is to predict a set of strokes which can cover the differences between a intermediate canvas image and a target image.
Meanwhile, to achieve a certain degree of abstraction to simulate real painting process, we hope the Stroke Predictor can predict as few strokes, while still covering most areas of differences. 
To achieve this, inspired by \textit{DETR}~\cite{carion2020endtoend}, we propose a Transformer-based predictor which takes in $I_c$ and $I_t$ and generates a stroke set, i.e.,

\begin{equation}
S_r = StrokePredictor(I_c, I_t).
\end{equation}

\begin{figure}[t]
\begin{center}
\includegraphics[width=0.9\linewidth]{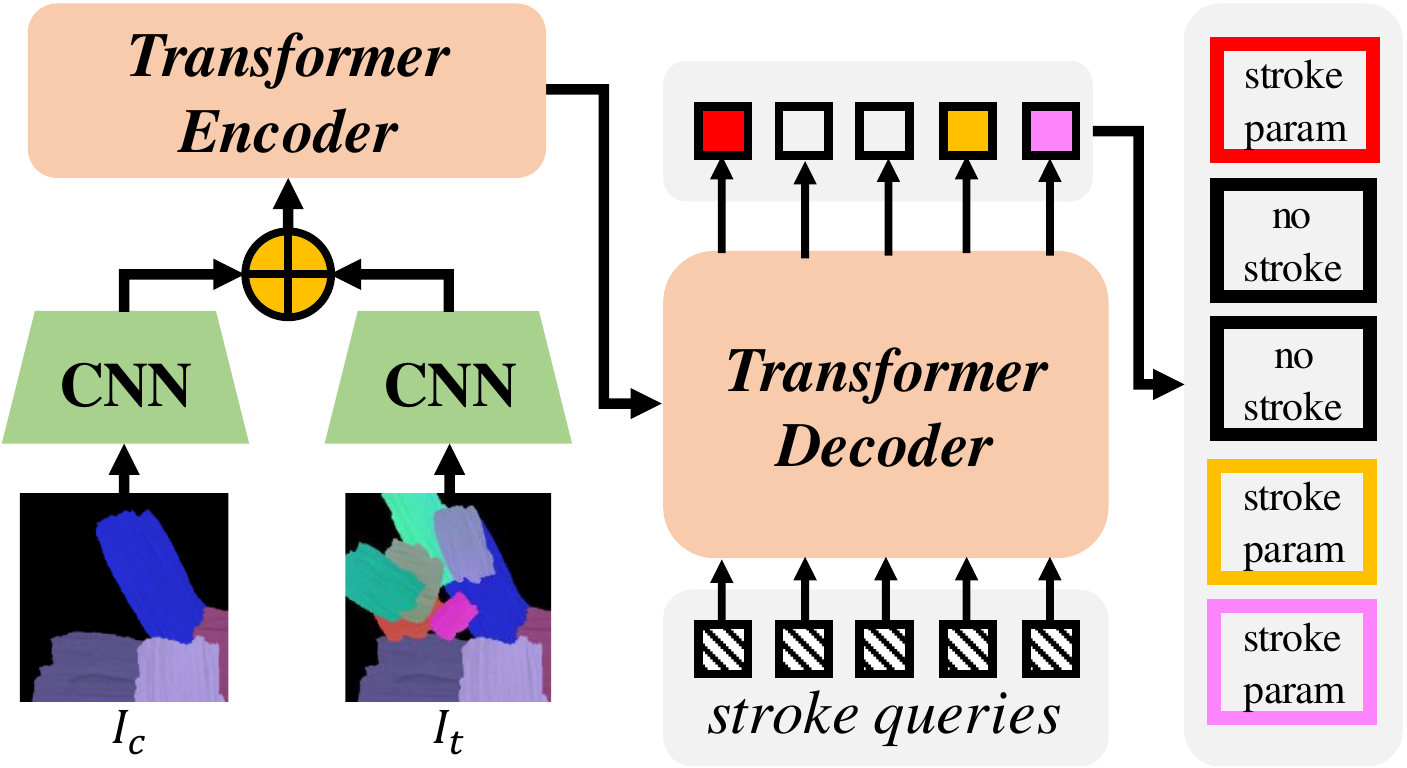}
\end{center}
   \caption{Illustration of Stroke Predictor, which contains two convolution networks for feature embedding and a Transformer network for stroke parameter prediction. $\oplus $ stands for concatenate.}
\label{fig:stroke_predictor}
\vspace{-0.5cm}
\end{figure}

As shown in Fig.~\ref{fig:stroke_predictor}, taking $I_c, I_t \in R^{3\times P \times P}$ as input, Stroke Predictor first adopts two independent convolution neural networks to extract their feature maps as $F_c, F_t \in R^{C\times P/4 \times P/4}$.
Here, $P$ is the pre-defined size of stroke image. Then, $F_c$, $F_t$ and a learnable positional encoding are concatenated and flattened as the input of Transformer encoder. 
In decoder part, following \textit{DETR}, we use $N$ learnable stroke query vectors as input.
Finally, there are two branches of fully-connected layers to predict initial stroke parameters $\bar{S}_r = \left \{ s_i \right \}_{i=1}^N$ and stroke confidence $C_r = \left \{ c_i \right \}_{i=1}^N$ respectively. 
Here, we add binary neurons for stroke confidence: in forward phase, confidence score $c_i$ can be converted to a decision $d_i = Sign(c_i)$,
%
where \textit{Sign} is a binary function, whose value is 1 if $c_i \geq 0$ and is 0 otherwise.
The decision $d_i$ is used to determine whether a predicted stroke should be plotted in canvas. 
Note that \textit{Sign} function has zero gradient almost everywhere. In order to enable back propagation, in backward phase, we alternatively utilize \textit{Sigmoid} function $\sigma(x)$ to compute gradient as:

\begin{equation}
    \frac{\partial d_i}{\partial c_i}=\frac{\partial\sigma(c_i)}{\partial c_i}=\frac{\exp(-c_i)}{(1+\exp(-c_i))^2}.
\end{equation}
Gathering all predicted strokes with positive decisions, we can get the final $S_r = \left \{ s_i \right \}_{i=1}^n$ with $n$ strokes.

%



\subsection{Loss Function}\label{sec_loss}

The major advantage of our proposed self-training pipeline is that we can simultaneously minimize differences between ground truth and prediction on both image level and stroke level. In this section, we will introduce our pixel loss, measurement of differences between strokes, and stroke loss. 


\noindent
\textbf{Pixel Loss.}
%
One intuitive goal for neural painting is to recreate a target image.
Therefore, pixel-wise loss $\mathcal{L}_{pixel}$ between $I_r$ and $I_t$ is penalized on the image level:
\begin{equation}
    \mathcal{L}_{pixel}=\left|\left|I_r-I_t\right|\right|_1. \label{l_pixel}
\end{equation}

\noindent
\textbf{Stroke Distance.}
On the stroke level, it is important to define appropriate metrics for measuring the difference between two strokes.
First, similar to the object detection task, we define parameter-wise $L_1$ distance as:
\begin{equation}
    \mathcal{D}^{u,v}_{L_1}=\left|\left|s_u-s_v\right|\right|_1,\label{l_gt}
\end{equation}
where $s_u$ and $s_v$ denote parameters of strokes $u$ and $v$ respectively.
As shown in many object detection works, merely employing the $L_1$ metric dismisses different scales for big and small strokes.
Thus, we further add the Wasserstein distance between two strokes following the idea in rotational object detection~\cite{yang2021rethinking}.
To be specific, a rotational rectangular stroke with parameters $[x,y,w,h,\theta]$ (excluding color parameters) can be viewed as a 2-D Gaussian distribution $\mathcal{N}(\mathbf{\mu},\mathbf{\Sigma})$ by the following equations:
\begin{equation}
\begin{aligned}
    \mathbf{\mu}&=(x,y),\\
    \mathbf{\Sigma}^{\frac{1}{2}}
    &=\begin{bmatrix}
        \cos\theta & -\sin\theta \\
        \sin\theta & \cos\theta \\
    \end{bmatrix}
    \begin{bmatrix}
        \frac{w}{2} & 0 \\
        0 & \frac{h}{2} \\
    \end{bmatrix}
    \begin{bmatrix}
        \cos\theta & \sin\theta \\
        -\sin\theta & \cos\theta \\
    \end{bmatrix}\\
    &=\begin{bmatrix}
        \frac{w}{2}\cos^2\theta+\frac{h}{2}\sin^2\theta & \frac{w-h}{2}\cos\theta\sin\theta \\
        \frac{w-h}{2}\cos\theta\sin\theta & \frac{w}{2}\sin^2\theta+\frac{h}{2}\cos^2\theta \\
    \end{bmatrix}.
\end{aligned}
\end{equation}

Therefore, the Wasserstein distance between two Gaussian distributions $\mathcal{N}(\mathbf{\mu}_u,\mathbf{\Sigma}_u)$ and $\mathcal{N}(\mathbf{\mu}_v,\mathbf{\Sigma}_v)$ is:
\begin{equation}
    \mathcal{D}^{{u,v}}_{W}=\left|\left|\mathbf{\mu}_u-\mathbf{\mu}_v\right|\right|_2^2+Tr(\mathbf{\Sigma}_u+\mathbf{\Sigma}_v-2(\mathbf{\Sigma}_u^{\frac{1}{2}}\mathbf{\Sigma}_v\mathbf{\Sigma}_u^{\frac{1}{2}})^{\frac{1}{2}}),\label{l_w}
\end{equation}
where $Tr(\cdot)$ denotes the trace of a matrix.
Moreover, it is desired that the predicted confidence for a stroke with positive (negative) ground-truth decision should be as high (low) as possible.
Let's consider $s_u$ as a predicted stroke with confidence $c_u$ and $s_v$ as a target stroke with ground-truth label $g_v$, where $g_v = 1$ if $s_v$ is a valid stroke and $g_v = 0$ if $s_v$ is an empty stroke.
Therefore, we can utilize binary cross entropy to match the confidence similarity:

\begin{equation}
    \mathcal{D}^{u,v}_{bce}=-\lambda_r \cdot g_v\cdot \log\sigma(c_u)-(1-g_v)\cdot \log (1-\sigma(c_u)),
\end{equation}
where $\lambda_r$ is a weight term controlling recall.

\noindent
\textbf{Stroke Loss.}
During training, the number of valid ground-truth strokes is varied. Thus, following DETR~\cite{carion2020endtoend}, with the predefined maximum stroke number $N$, we need to first generate a matching mechanism between the prediction set $\bar{S}_r$ of $N$ strokes and the ground-truth set $S_g$ of $N$ strokes (they can be both valid and empty strokes in $S_g$) to calculate the loss. 
%
Following DETR~\cite{carion2020endtoend}, we adopt the permutation of strokes that produces the minimal stroke level matching cost to calculate final loss. 
The optimal bipartite matching is firstly computed leveraging the Hungarian algorithm~\cite{kuhn1955hungarian}.
For a stroke $s_u$ in the prediction set $\bar{S}_r$ and a stroke $s_v$ in the target set $S_g$, their cost value is:

\begin{equation}
    M_{u, v}=g_v(\mathcal{D}^{u,v}_{L_1}+\mathcal{D}^{u,v}_W+\mathcal{D}^{u,v}_{bce}),
\end{equation}
which means the matching cost for empty target strokes is always $0$.
Therefore, denoting as $X$ and $Y$ the optimal permutations for predicted strokes and target strokes given by the Hungarian algorithm, respectively, the stroke loss function can be written as:
\begin{equation}
\begin{aligned}
    \mathcal{L}_{stroke}=\frac{1}{n}\sum_{i=1}^n&(g_{Y_i}(\lambda_{L_1}\mathcal{D}^{X_iY_i}_{L_1}+\lambda_{W}\mathcal{D}^{X_iY_i}_{W})\\&+\lambda_{bce}\mathcal{D}^{X_iY_i}_{bce}),
\end{aligned}
\end{equation}
where $\lambda_{L_1}$, $\lambda_W$, and $\lambda_{bce}$ are weight terms.
Moreover, although in the neural painting task, stroke order is of great importance, we ignore the stroke order in the stroke level loss and set the task of regulating stroke order to the image level.


\begin{algorithm}[!t]
	\caption{Inference Algorithm of Paint Transformer}
    \textbf{Required}: A target image $I_t$ with shape $H\times W$; Stroke Predictor $SP$; Stroke Renderer $SR$.
    \begin{algorithmic}[1]
        \State $K=\max(\mathrm{argmin}_K\{P\times 2^K\geq\max(H,W)\},0)$;
        \State $I_t=pad(I_t,size=(P\times 2^K,P\times 2^K))$;
        \State $I_c=blank\_canvas$;
        \For{$0\leq k \leq K$}
            \State $I_t^k=resize(I_t,(P\times 2^k,P\times 2^k))$;
            \State $I_c^k=resize(I_c,(P\times 2^k,P\times 2^k))$;
            \State $I_t^{k^{'}}=image\_to\_patches(I_t^k,size=(P,P))$;
            \State $I_c^{k^{'}}=image\_to\_patches(I_c^k,size=(P,P))$;
            \State $S_r^{k}=SP(I_t^{k^{'}},I_c^{k^{'}})$;
            \State $I_r^{k}=SR(S_r^{k}, I_c^{k^{'}})$;
            \State $I_c=patches\_to\_image(I_r^{k})$;
        \EndFor
        \State $I_r=crop(I_c,size=(H,W))$;\\
        \Return $I_r$.
	\end{algorithmic}
	\label{inference}
	
\end{algorithm}

\begin{figure*}[t]
\begin{center}
\includegraphics[width=0.98\linewidth]{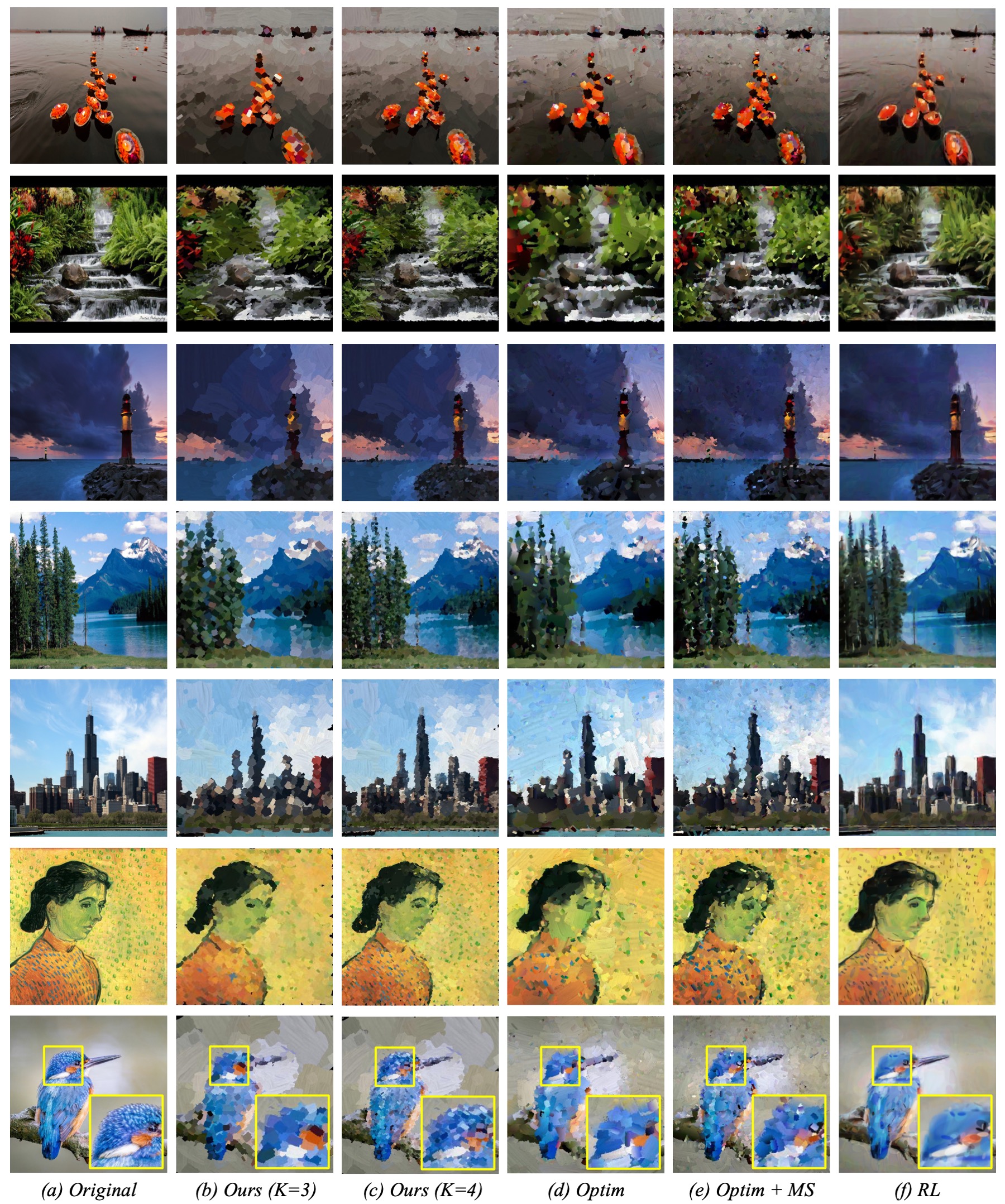}
\end{center}
   \caption{Comparison with the state-of-the-art methods: an optimization-based method (\emph{Optim}) \cite{zou2020stylized} and an RL-based method \cite{huang2019learning}. We also demonstrate our results with different number of rendering scales, where $K=4$ is the default setting. \emph{MS} here denotes using more strokes for \emph{Optim}, with same number as \emph{$Ours~(K=4)$} .}
\label{fig:comparison_optim}
\end{figure*}

\subsection{Inference}
To imitate a human painter, we devise a coarse-to-fine algorithm to generate painting results during inference, as shown in Algorithm~\ref{inference}. 
Given a real-world image of size $H\times W$, our Paint Transformer runs on $K$ scales from coarse to fine in order.
Painting on each scale is dependent on result of the previous scale. 
Target image and current canvas would be cut into several non-overlapping $P\times P$ patches before being sent to Stroke Predictor.
We set $K$ as follow:
\begin{equation}
    K=\max(\mathrm{argmin}_K\{P\times 2^K\geq\max(H,W)\},0),
\end{equation}
where in the $k$-th ($0\leq k \leq K$) scale, there are $2^k\times 2^k$ patches. 
Each patch would go through Stroke Predictor and then Stroke Renderer module in parallel independently. 
The painting result on each scale is derived by combining patches of canvas together. 

\begin{figure*}[t]
\begin{center}
\includegraphics[width=0.99\linewidth]{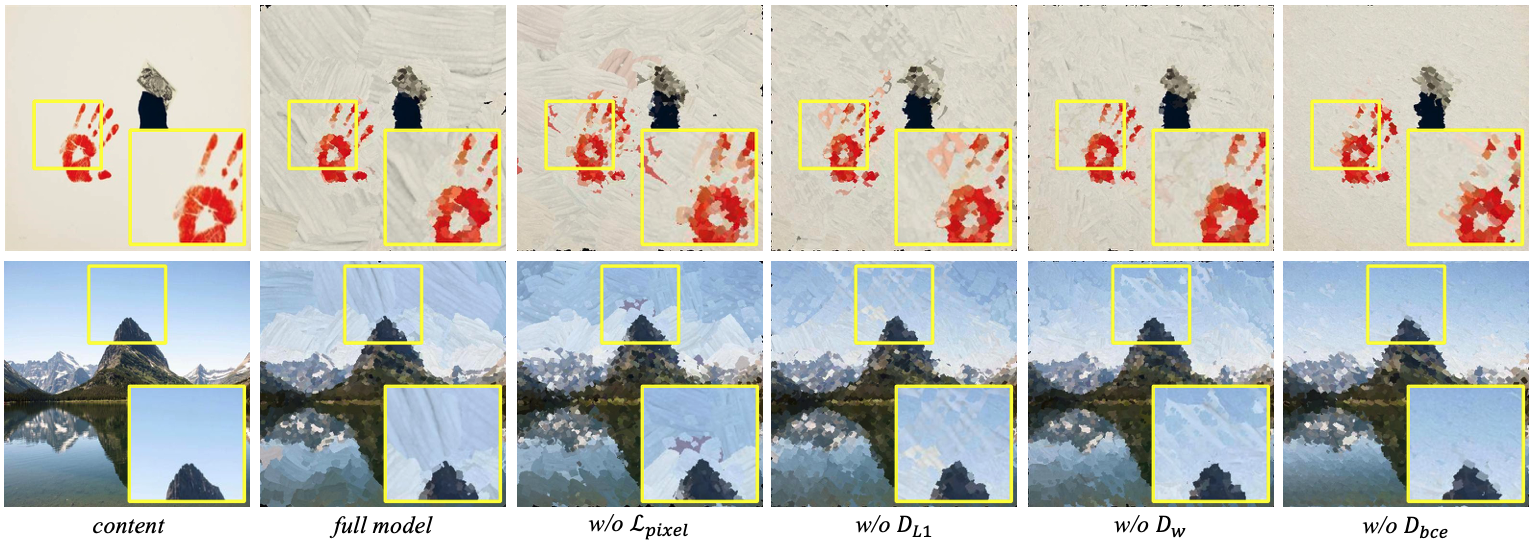}
\end{center}
\vspace{-0.5cm}
   \caption{Ablation study on proposed different loss terms. To illustrate the differences clearly, in each image, an area is enlarged.}
   \vspace{-0.2cm}
\label{fig:exp_ablation}
\end{figure*}

\section{Experiments}

\subsection{Implementing Details}
To train our Paint Transformer, in practice, we set the size of input images $P$ as $32$, and the number of strokes $N$ in one patch as $8$.
The CNNs for image feature extraction consists of three Conv-BatchNorm-ReLU blocks with two $1/2$-scale down-sampling operations.
For the Transformer, the feature dimension is $256$ and both encoder and decoder have $3$ layers.
During training, we randomly generate parameters of $N$ target strokes from a uniform distribution.
To prevent too much stroke-wise overlap and ensure that the number of valid target strokes is varied, we generate strokes for $I_t$ in sequence and set the label of a stroke to $0$ if it covers more than 60\% area of one previous stroke.
Hyper-parameters $\lambda_r$, $\lambda_{L1}$, $\lambda_w$ and $\lambda_{bce}$ are set to $8$, $1$, $10$, and $1$ respectively.
We use the \textit{Adam} optimizer~\cite{kingma2014adam} with a learning rate of $0.0001$. 
We train the model for $30,000$ iterations with a batch size of $128$ on a single Nvidia RTX 2080 Ti GPU. The total training time is fewer than $4$ hours.
For inference, painting results in this paper are all under $512\times 512$ resolution with $K=4$ if not specified.

\begin{table}[!t]
\centering
    \begin{tabular}{lccccc}
        \toprule
        \multirow{2}{1.5cm}{Methods} & \multicolumn{2}{c}{Real Images} & \multicolumn{3}{c}{Random Strokes} \\
        \cline{2-6}
        & $\mathcal{L}_{pixel}$ & $\mathcal{L}_{pcpt}$ & $\mathcal{L}_{pixel}$ & $D_{L_1}$ & $D_W$ \\
        \midrule
        RL~\cite{huang2019learning} & \textbf{0.040} & \textbf{0.737} & 0.058 & - & - \\
        Optim~\cite{zou2020stylized} & 0.059 & 0.856 & 0.073 & 0.137 & 0.057 \\
        Ours & 0.056 & 0.807 & \textbf{0.042} & \textbf{0.083} & \textbf{0.018} \\
        \hline
        \textit{w/o} $\mathcal{L}_{pixel}$ & 0.081 & 1.012 & 0.068 & 0.241 & 0.024 \\
        \textit{w/o} $\mathcal{D}_{L_1}$ & 0.074 & 0.941 & 0.077 & 0.267 & 0.019 \\
        \textit{w/o} $\mathcal{D}_{W}$ & 0.069 & 0.947 & 0.046 & 0.113 & 0.034 \\
        \textit{w/o} $\mathcal{D}_{bce}$ & 0.071 & 0.928 & 0.052 & 0.093 & 0.021 \\
        \bottomrule
    \end{tabular}
    \vspace{0.3cm}
    \caption{Quantitative results under different metrics for different methods or settings. Smaller values mean closer to original inputs. \emph{Optim} is applied with the same number of strokes with Ours.}
    \vspace{-0.5cm}
    \label{table:quantitative}
\end{table}

\subsection{Comparison with State-of-the-Art Methods}

\noindent
\textbf{Qualitative Comparison.}
As shown in Fig.~\ref{fig:comparison_optim}, we compare our method with two state-of-the-art stroke-based painting generation methods.
Comparing with the optimization-based method (\emph{Optim}) \cite{zou2020stylized}, our method can generate more appealing and refreshing results. To be specific, in large texture-less image areas, our method can generate human-like painting with relative fewer and bigger strokes (row 3, 5 and 7). In small texture-rich image areas, our method can generate painting with clearer texture to preserve content structure. We further implement \emph{Optim} with more strokes (column 5), however, the aforementioned problem still exists.
Compared with the RL-based method~\cite{huang2019learning}, we can generate more vivid results with clear brushes. Meanwhile, the results of \cite{huang2019learning} are somehow blurred and lack of artistic abstraction, it is also too similar to the original images. 

\begin{table}[!t]
\centering
    \begin{tabular}{lcccc}
        \toprule
        \multicolumn{2}{c}{Method} & Ours & RL~\cite{huang2019learning} & Optim~\cite{zou2020stylized} \\
        \midrule
        \multirow{3}{2cm}{Inference (s)} & 128 & \textbf{0.055} & 0.242 & 76.508 \\
        \cline{2-5}
        & 256 & \textbf{0.124} & 0.266 & 161.591 \\
        \cline{2-5}
        & 512 & \textbf{0.304} & 0.322 & 521.447 \\
        \hline
        \multirow{2}{2cm}{Training (h)} & $SP$ & 3.79 & 40 & 0 \\
        \cline{2-5}
        & $SR$ & 0 & 5-15 & 10.16 \\
        \hline
        \multicolumn{2}{c}{Use External Dataset} & No & Yes & No \\
        \bottomrule
    \end{tabular}
    \vspace{0.3cm}
    \caption{Efficiency of inference and training for different methods.}
    \label{table:efficiency}
    \vspace{-0.2cm}
\end{table}

\noindent
\textbf{Quantitative Comparison.}
We also conduct quantitative comparison for reference. 
Since one objective of neural painting is to recreate original images, we directly use the pixel loss $\mathcal{L}_{pixel}$ and the perceptual loss $\mathcal{L}_{pcpt}$ \cite{johnson2016perceptual} as evaluation metrics.
For real images, we randomly select $100$ landscapes from \cite{arnaud58}, $100$ artworks from \textit{WikiArt}~\cite{phillips2011wiki}, and $100$ portraits from FFHQ \cite{karras2019stylebased} for evaluation. 
Results shown in Table~\ref{table:quantitative} are consistent with the previous qualitative analysis: (1) with vivid brush textures, our method can present the original content better than \emph{Optim}~\cite{zou2020stylized}; (2) \cite{huang2019learning} achieves the best content fidelity, however it is weak in abstraction.
Then, to compare stroke prediction performance, we send synthesized stroke images to both \emph{Paint Transformer} and \emph{Optim} and evaluate their generated strokes with the same metrics as Sec.~\ref{sec_loss}.
Numeric results show that our method can predict strokes successfully and outperforms other methods. 
Here, measurements are missing for \cite{huang2019learning}, since it has differently parameterized strokes.

\noindent
\textbf{Efficiency Analysis.}
We demonstrate efficiency comparison in Table \ref{table:efficiency}.
Training or inference time is measured using a single Nvidia 2080Ti GPU.
During inference, since Paint Transformer produces a set of strokes in parallel in a feed-forward manner, it runs significantly faster than optimization baseline~\cite{zou2020stylized} and slightly faster than the RL-based baseline \cite{huang2019learning}.
As for training, we only need a few hours to train a Stroke Predictor, which is more convenient than both \cite{huang2019learning} and \cite{zou2020stylized} from the perspective of total training time.
Besides, our model-free Stroke Renderer and data-free Stroke Predictor are efficient and convenient to use.

\begin{figure}[t]
\begin{center}
\includegraphics[width=0.99\linewidth]{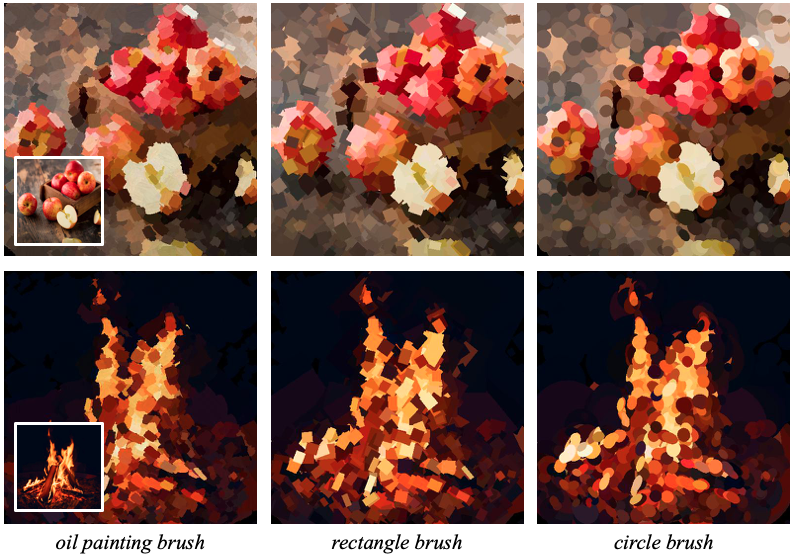}
\end{center}
\vspace{-0.5cm}
   \caption{Results with different brushes, $K$ is set to 3 here.}
\label{fig:change_stroke}
\vspace{-0.5cm}
\end{figure}


\subsection{Ablation study}

As shown in Fig.~\ref{fig:exp_ablation}, we present ablation study results to verify the effectiveness of each optimization term used for training Paint Transformer. 
\textbf{(1)} Without the pixel loss, the model fails to learn proper locations to put strokes with proper colors, resulting in wrong colors and dirty textures;
\textbf{(2)} Without the parameter $L_1$ loss, the model fails to learn the shapes of strokes and present repeated stroke patterns;
\textbf{(3)} Without the Wasserstein loss, it seems that the ability of handling strokes with different scales is weakened, with large and vivid strokes vanished;
\textbf{(4)} Without the confidence loss, the model cannot decide whether to plot a stroke or not, resulting in too many small strokes totally covering the whole image and previous strokes. 
We also present quantitative ablation results in Table \ref{table:quantitative}, which demonstrates that missing each of proposed metric leads to performance drop.

\subsection{Extension of Paint Transformer}

\begin{figure}[t]
\begin{center}
\includegraphics[width=0.99\linewidth]{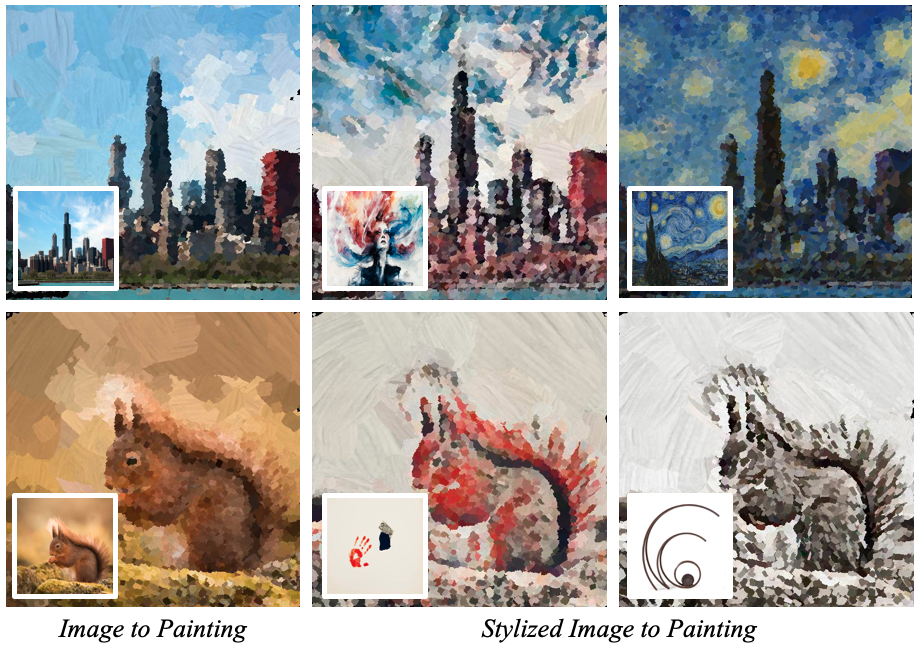}
\end{center}
\vspace{-0.2cm}
   \caption{Stylized paintings.}
\label{fig:exp_style}
\vspace{-0.5cm}
\end{figure}

\noindent
\textbf{Painting with Different Strokes. }
Notably, once trained with one kind of primitive brush, our Stroke Predictor can be easily transferred to another kind via replacing the primitive brush used in Stroke Renderer.
As demonstrated in Fig.~\ref{fig:change_stroke}, with a Stroke Predictor trained with oil-painting brush, we can still generate appealing and vivid painting results with rectangle and circle brushes.

\noindent
\textbf{Stylized Painting. }
%
%
It is also flexible for our method to be integrated with artistic style transfer to generate attractive and stylized paintings. 
We utilize existing style transfer methods such as LapStyle~\cite{lin2021drafting} and AdaAttN~\cite{liu2021adaattn} to generate neural paintings on stylized content images.
As shown in Fig.~\ref{fig:exp_style}, with this imaginative manner, we can generate stylized paintings with diverse colors and textures.

\section{Conclusion and Future Works}
In this paper, we re-formulate the neural painting problem from a stroke set prediction perspective.
Leveraging insights from Transformer-based object detection, we propose a novel framework, dubbed Paint Transformer, which can generate paintings from natural images via predicting parameters of multiple strokes with a feed-forward Transformer.
Moreover, we propose a novel self-training pipeline that makes it possible to train our Paint Transformer without any manually collected dataset. 
Experiments demonstrate that our model can generate paintings with better trade-off between artistic abstraction and realism, compared to state-of-the-art methods, while maintaining high efficiency. 

As for our future work, it is a valuable topic to explore more complex strokes with various shapes or color patterns besides straight-line strokes with uniform colors. 
More advanced stroke rendering systems are required for these stroke settings. 
It may further improve the painting quality of long but narrow areas if cross-patch context is exploited. 



{\small
\bibliographystyle{ieee_fullname}
\balance
\bibliography{NeuralPainting}

\begin{thebibliography}{10}\itemsep=-1pt

\bibitem{arnaud58}
Datasets of pictures of natural landscapes.
\newblock Website, 2020.
\newblock \url{https://www.kaggle.com/arnaud58/landscape-pictures}.

\bibitem{carion2020endtoend}
Nicolas Carion, Francisco Massa, Gabriel Synnaeve, Nicolas Usunier, Alexander
  Kirillov, and Sergey Zagoruyko.
\newblock End-to-end object detection with transformers, 2020.

\bibitem{chen2018cartoongan}
Yang Chen, Yu-Kun Lai, and Yong-Jin Liu.
\newblock Cartoongan: Generative adversarial networks for photo cartoonization.
\newblock In {\em Proceedings of the IEEE conference on computer vision and
  pattern recognition}, pages 9465--9474, 2018.

\bibitem{ganin2018synthesizing}
Yaroslav Ganin, Tejas Kulkarni, Igor Babuschkin, SM~Ali Eslami, and Oriol
  Vinyals.
\newblock Synthesizing programs for images using reinforced adversarial
  learning.
\newblock In {\em International Conference on Machine Learning}, pages
  1666--1675. PMLR, 2018.

\bibitem{gatys2016image}
Leon~A Gatys, Alexander~S Ecker, and Matthias Bethge.
\newblock Image style transfer using convolutional neural networks.
\newblock In {\em Proceedings of the IEEE conference on computer vision and
  pattern recognition}, pages 2414--2423, 2016.

\bibitem{ha2017neural}
David Ha and Douglas Eck.
\newblock A neural representation of sketch drawings.
\newblock {\em arXiv preprint arXiv:1704.03477}, 2017.

\bibitem{haeberli1990paint}
Paul Haeberli.
\newblock Paint by numbers: Abstract image representations.
\newblock In {\em Proceedings of the 17th annual conference on Computer
  graphics and interactive techniques}, pages 207--214, 1990.

\bibitem{hertzmann1998painterly}
Aaron Hertzmann.
\newblock Painterly rendering with curved brush strokes of multiple sizes.
\newblock In {\em Proceedings of the 25th annual conference on Computer
  graphics and interactive techniques}, pages 453--460, 1998.

\bibitem{huang2017arbitrary}
Xun Huang and Serge Belongie.
\newblock Arbitrary style transfer in real-time with adaptive instance
  normalization.
\newblock In {\em Proceedings of the IEEE International Conference on Computer
  Vision}, pages 1501--1510, 2017.

\bibitem{huang2019learning}
Zhewei Huang, Wen Heng, and Shuchang Zhou.
\newblock Learning to paint with model-based deep reinforcement learning.
\newblock In {\em Proceedings of the IEEE/CVF International Conference on
  Computer Vision}, pages 8709--8718, 2019.

\bibitem{isola2017image}
Phillip Isola, Jun-Yan Zhu, Tinghui Zhou, and Alexei~A Efros.
\newblock Image-to-image translation with conditional adversarial networks.
\newblock In {\em Computer Vision and Pattern Recognition (CVPR), 2017 IEEE
  Conference on}, 2017.

\bibitem{johnson2016perceptual}
Justin Johnson, Alexandre Alahi, and Li Fei-Fei.
\newblock Perceptual losses for real-time style transfer and super-resolution.
\newblock In {\em European conference on computer vision}, pages 694--711.
  Springer, 2016.

\bibitem{karras2019stylebased}
Tero Karras, Samuli Laine, and Timo Aila.
\newblock A style-based generator architecture for generative adversarial
  networks, 2019.

\bibitem{kato2020differentiable}
Hiroharu Kato, Deniz Beker, Mihai Morariu, Takahiro Ando, Toru Matsuoka, Wadim
  Kehl, and Adrien Gaidon.
\newblock Differentiable rendering: A survey.
\newblock {\em arXiv preprint arXiv:2006.12057}, 2020.

\bibitem{kingma2014adam}
Diederik~P Kingma and Jimmy Ba.
\newblock Adam: A method for stochastic optimization.
\newblock {\em arXiv preprint arXiv:1412.6980}, 2014.

\bibitem{kolkin2019style}
Nicholas Kolkin, Jason Salavon, and Gregory Shakhnarovich.
\newblock Style transfer by relaxed optimal transport and self-similarity.
\newblock In {\em Proceedings of the IEEE/CVF Conference on Computer Vision and
  Pattern Recognition}, pages 10051--10060, 2019.

\bibitem{kotovenko_cvpr_2021}
Dmytro Kotovenko, Matthias Wright, Arthur Heimbrecht, and Bj{\"o}rn Ommer.
\newblock Rethinking style transfer: From pixels to parameterized brushstrokes.
\newblock {\em CVPR}, 2021.

\bibitem{kuhn1955hungarian}
Harold~W Kuhn.
\newblock The hungarian method for the assignment problem.
\newblock {\em Naval research logistics quarterly}, 2(1-2):83--97, 1955.

\bibitem{li2018learning}
Xueting Li, Sifei Liu, Jan Kautz, and Ming-Hsuan Yang.
\newblock Learning linear transformations for fast arbitrary style transfer.
\newblock {\em arXiv preprint arXiv:1808.04537}, 2018.

\bibitem{lin2021drafting}
Tianwei Lin, Zhuoqi Ma, Fu Li, Dongliang He, Xin Li, Errui Ding, Nannan Wang,
  Jie Li, and Xinbo Gao.
\newblock Drafting and revision: Laplacian pyramid network for fast
  high-quality artistic style transfer.
\newblock In {\em Proceedings of the IEEE/CVF Conference on Computer Vision and
  Pattern Recognition}, pages 5141--5150, 2021.

\bibitem{litwinowicz1997processing}
Peter Litwinowicz.
\newblock Processing images and video for an impressionist effect.
\newblock In {\em Proceedings of the 24th annual conference on Computer
  graphics and interactive techniques}, pages 407--414, 1997.

\bibitem{liu2021adaattn}
Songhua Liu, Tianwei Lin, Dongliang He, Fu Li, Meiling Wang, Xin Li, Zhengxing
  Sun, Qian Li, and Errui Ding.
\newblock Adaattn: Revisit attention mechanism in arbitrary neural style
  transfer.
\newblock In {\em Proceedings of the IEEE International Conference on Computer
  Vision}, 2021.

\bibitem{nakano2019neural}
Reiichiro Nakano.
\newblock Neural painters: A learned differentiable constraint for generating
  brushstroke paintings.
\newblock {\em arXiv preprint arXiv:1904.08410}, 2019.

\bibitem{park2019arbitrary}
Dae~Young Park and Kwang~Hee Lee.
\newblock Arbitrary style transfer with style-attentional networks.
\newblock In {\em Proceedings of the IEEE/CVF Conference on Computer Vision and
  Pattern Recognition}, pages 5880--5888, 2019.

\bibitem{phillips2011wiki}
Fred Phillips and Brandy Mackintosh.
\newblock Wiki art gallery, inc.: A case for critical thinking.
\newblock {\em Issues in Accounting Education}, 26(3):593--608, 2011.

\bibitem{yolo}
Joseph Redmon, Santosh Divvala, Ross Girshick, and Ali Farhadi.
\newblock You only look once: Unified, real-time object detection.
\newblock In {\em Proceedings of the IEEE conference on computer vision and
  pattern recognition}, pages 779--788, 2016.

\bibitem{redmon2016look}
Joseph Redmon, Santosh Divvala, Ross Girshick, and Ali Farhadi.
\newblock You only look once: Unified, real-time object detection, 2016.

\bibitem{ren2015faster}
Shaoqing Ren, Kaiming He, Ross Girshick, and Jian Sun.
\newblock Faster r-cnn: Towards real-time object detection with region proposal
  networks.
\newblock {\em arXiv preprint arXiv:1506.01497}, 2015.

\bibitem{tian2019fcos}
Zhi Tian, Chunhua Shen, Hao Chen, and Tong He.
\newblock Fcos: Fully convolutional one-stage object detection, 2019.

\bibitem{vaswani2017attention}
Ashish Vaswani, Noam Shazeer, Niki Parmar, Jakob Uszkoreit, Llion Jones,
  Aidan~N. Gomez, Lukasz Kaiser, and Illia Polosukhin.
\newblock Attention is all you need, 2017.

\bibitem{wang2020learning}
Xinrui Wang and Jinze Yu.
\newblock Learning to cartoonize using white-box cartoon representations.
\newblock In {\em Proceedings of the IEEE/CVF Conference on Computer Vision and
  Pattern Recognition}, pages 8090--8099, 2020.

\bibitem{xie2013artist}
Ning Xie, Hirotaka Hachiya, and Masashi Sugiyama.
\newblock Artist agent: A reinforcement learning approach to automatic stroke
  generation in oriental ink painting.
\newblock {\em IEICE TRANSACTIONS on Information and Systems},
  96(5):1134--1144, 2013.

\bibitem{yang2021rethinking}
Xue Yang, Junchi Yan, Qi Ming, Wentao Wang, Xiaopeng Zhang, and Qi Tian.
\newblock Rethinking rotated object detection with gaussian wasserstein
  distance loss, 2021.

\bibitem{yi2019apdrawinggan}
Ran Yi, Yong-Jin Liu, Yu-Kun Lai, and Paul~L Rosin.
\newblock Apdrawinggan: Generating artistic portrait drawings from face photos
  with hierarchical gans.
\newblock In {\em Proceedings of the IEEE/CVF Conference on Computer Vision and
  Pattern Recognition}, pages 10743--10752, 2019.

\bibitem{yi2020unpaired}
Ran Yi, Yong-Jin Liu, Yu-Kun Lai, and Paul~L Rosin.
\newblock Unpaired portrait drawing generation via asymmetric cycle mapping.
\newblock In {\em Proceedings of the IEEE/CVF Conference on Computer Vision and
  Pattern Recognition}, pages 8217--8225, 2020.

\bibitem{zheng2018strokenet}
Ningyuan Zheng, Yifan Jiang, and Dingjiang Huang.
\newblock Strokenet: A neural painting environment.
\newblock In {\em International Conference on Learning Representations}, 2018.

\bibitem{zhou2018learning}
Tao Zhou, Chen Fang, Zhaowen Wang, Jimei Yang, Byungmoon Kim, Zhili Chen,
  Jonathan Brandt, and Demetri Terzopoulos.
\newblock Learning to sketch with deep q networks and demonstrated strokes.
\newblock {\em arXiv preprint arXiv:1810.05977}, 2018.

\bibitem{CycleGAN2017}
Jun-Yan Zhu, Taesung Park, Phillip Isola, and Alexei~A Efros.
\newblock Unpaired image-to-image translation using cycle-consistent
  adversarial networkss.
\newblock In {\em Computer Vision (ICCV), 2017 IEEE International Conference
  on}, 2017.

\bibitem{zou2020stylized}
Zhengxia Zou, Tianyang Shi, Shuang Qiu, Yi Yuan, and Zhenwei Shi.
\newblock Stylized neural painting, 2020.

\end{thebibliography}
}

\appendix


\twocolumn[{%
\begin{figure}[H]
\hsize=\textwidth 
\centering
\includegraphics[width=\textwidth]{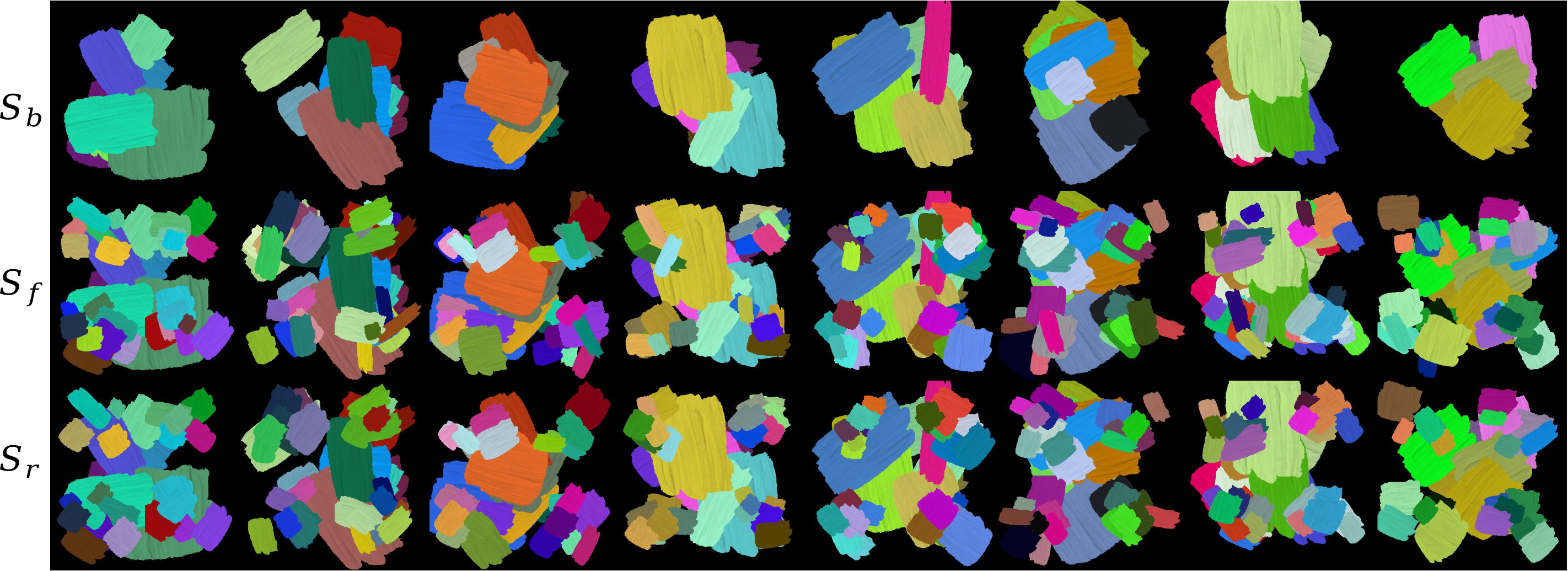}
\caption{Canvas-target-predict pairs in training.}
\label{fig_training_pairs}
\end{figure}
}]

\section{More Training Details}
In our training, firstly, we sample $8$ strokes on a canvas with $64\times64$ resolution as background. 
Then, the background canvas is divided into $4$ blocks with $32\times32$ size. 
For each block, we further sample another $8$ strokes as foreground ones based on the background. 
The stroke predictor learns to predict these extra strokes.
Such operations encourage the stroke predictor to paint from coarse to fine. 
Thus, it always tends to generate refined strokes to minimize the differences between current canvas and target. 
During inference, the coarse-to-fine inference process can gradually fill in the canvas and reduce the differences between canvas and the real image.
Therefore, our stroke predictor can be generalized from randomly-synthesized dataset to real-world images successfully. 
More canvas-target-predict pairs (denoted as $S_b$, $S_f$, and $S_r$ respectively) during training period are shown in Fig. \ref{fig_training_pairs}.

\section{More Inference Results}
\begin{figure*}[b]
\begin{center}
\includegraphics[width=0.62\textwidth]{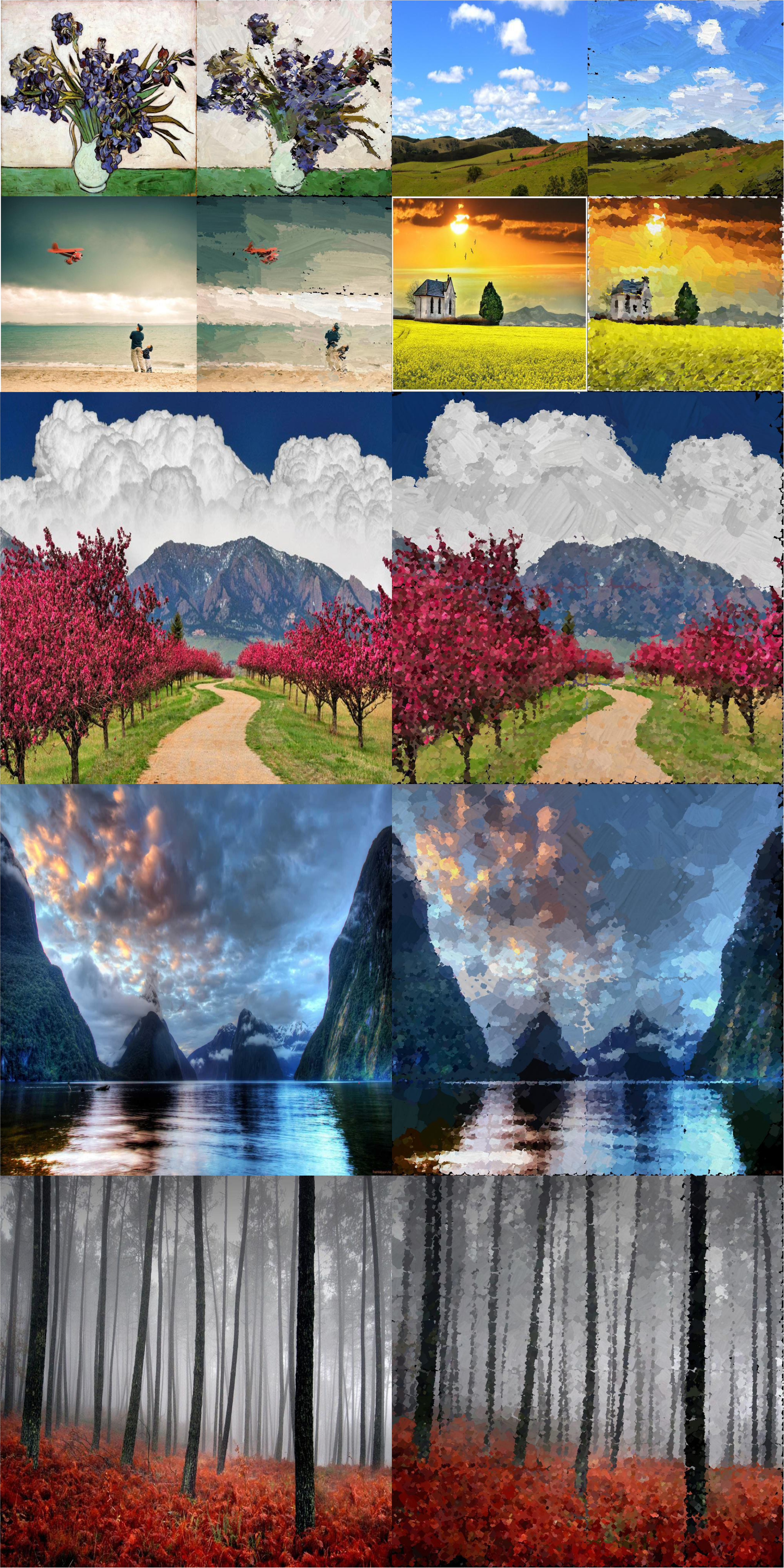}
\end{center}
\caption{More inference result.}
\label{fig_more_results}
\end{figure*}

We provide more results including high-resolution ($1024\times1024$) results of our algorithm in Fig. \ref{fig_more_results}. 
The animated painting process can be found in the attachment or \href{https://github.com/Huage001/PaintTransformer}{our code page}.

\end{document}